\documentclass{article}
\usepackage{spconf,amsmath,amssymb,epsfig}
\usepackage{hyperref}

\title{EMPIRICAL UPSCALING OF POINT-SCALE SOIL MOISTURE MEASUREMENTS FOR SPATIAL EVALUATION OF MODEL SIMULATIONS AND SATELLITE RETRIEVALS}

\name{
\begin{tabular}{c}Yi Yu$^{a,b,*}$, Brendan P. Malone$^{b}$, Luigi J. Renzullo$^{c}$\end{tabular}\thanks{$^{*}$Correspondence to Yi Yu (u6726739@anu.edu.au)}
}

\address{
$^{a}$ The Australian National University, Canberra, ACT 2601, Australia \\
$^{b}$ CSIRO Agriculture and Food, Canberra, ACT 2601, Australia \\
$^{c}$ Bureau of Meteorology, Canberra, ACT 2600, Australia
}

\begin{document}

\maketitle

\begin{abstract}
The evaluation of modelled or satellite-derived soil moisture (SM) estimates is usually dependent on comparisons against in-situ SM measurements. However, the inherent mismatch in spatial support (i.e., scale) necessitates a cautious interpretation of point-to-pixel comparisons. The upscaling of the in-situ measurements to a commensurate resolution to that of the modelled or retrieved SM will lead to a fairer comparison and statistically more defensible evaluation. In this study, we presented an upscaling approach that combines spatiotemporal fusion with machine learning to extrapolate point-scale SM measurements from 28 in-situ sites to a 100 m resolution for an agricultural area of 100 km × 100 km. We conducted a four-fold cross-validation, which consistently demonstrated comparable correlation performance across folds, ranging from 0.6 to 0.9. The proposed approach was further validated based on a cross-cluster strategy by using two spatial subsets within the study area, denoted as cluster A and B, each of which equally comprised of 12 in-situ sites. The cross-cluster validation underscored the capability of the upscaling approach to map the spatial variability of SM within areas that were not covered by in-situ sites, with correlation performance ranging between 0.6 and 0.8. In general, our proposed upscaling approach offers an avenue to extrapolate point measurements of SM to a spatial scale more akin to climatic model grids or remotely sensed observations. Future investigations should delve into a further evaluation of the upscaling approach using independent data, such as model simulations, satellite retrievals or field campaign data.
\end{abstract}


\section{Introduction}
Soil moisture (SM) is vital in global water, energy cycle and biogeochemical flux system \cite{SENEVIRATNE2010125}. Spatiotemporally continuous SM datasets have been extensively developed and explored throughout remote sensing techniques and model simulations \cite{rs10122038, DONG2020111756, PENG2021112162}. However, SM estimates commonly display discrepancies among different satellite platforms and models \cite{reichle2004global, koster2009nature}, even when subjected to identical meteorological forcings \cite{dirmeyer2004comparison, CHEN2020125054}. A thorough evaluation of these datasets usually depends on comparisons against in-situ SM measurements, while the inherent mismatch in spatial support (i.e., scale) necessitates a cautious interpretation of point-to-pixel comparisons.

At the agricultural scale, there is a necessity to enhance our comprehension and application of in-situ SM measurements. This knowledge is often accumulated and refined through the deployment of dense SM sensors. Existing in-situ SM monitoring networks and databases \cite{https://doi.org/10.2136/vzj2010.0139, https://doi.org/10.1029/2012WR011976, https://doi.org/10.1029/2011EO170001} always play a pivotal role in evaluating SM derived from model simulations or satellite retrievals. However, the representativeness of point measurements is constrained to a limited area and their distribution is often uneven \cite{BROCCA2007356, https://doi.org/10.1029/2011RG000372}, leading to gaps in coverage and potential biases in the validation process. Researchers have persistently raised questions about the effective alignment of SM footprints between in-situ data and grids from climatic models or satellite platforms, with upscaling techniques emerging as an appealing solution \cite{https://doi.org/10.1029/2011RG000372}. In this study, we presented an upscaling approach that integrates spatiotemporal fusion with machine learning to extrapolate point-scale SM measurements from 28 in-situ sites to a 100 m resolution. Subsequently, multiple iterations of validation were conducted to thoroughly assess the performance of the proposed approach.

\section{Study area and data}
\subsection{Study area and in-situ measurements}
\begin{figure}
    \centering
    \includegraphics[width=0.5\textwidth]{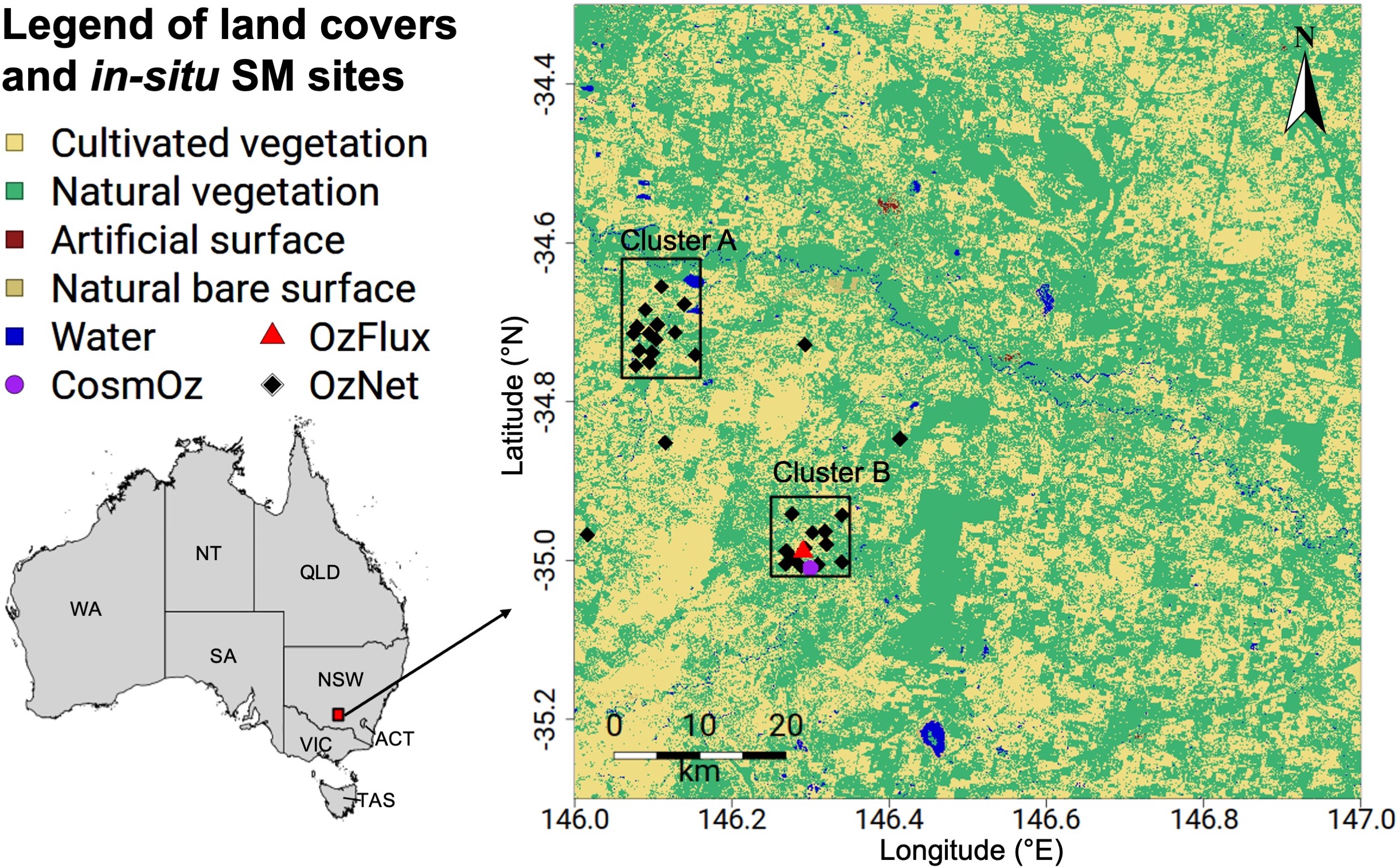}
    \caption{The location (highlighted in red rectangle within the NSW State) and the land cover information in 2020 of Yanco agricultural region. The purple circle, red triangle and black dots represent the locations of in-situ sties from CosmOz, OzFlux and OzNet, respectively. The cluster A spans the region of 146.06-146.16 °E and 34.62-34.77 °S (10 × 15 km\textsuperscript{2}) and the cluster B spans the region of 146.25-146.35 °E and 34.92-35.02 °S (10 × 10 km\textsuperscript{2}).}
    \label{fig1}
\end{figure}

We selected the Yanco agricultural region as the study area (Fig. \ref{fig1}). Yanco is situated within the Murrumbidgee Catchment, characterised by a semi-arid climate that is representative for most agricultural regions in southeast Australia. The annual total precipitation in this area averages approximately 400 mm \cite{SABAGHY2020111586}. We collected the in-situ SM data from the OzNet Hydrological Monitoring Network \cite{https://doi.org/10.1029/2012WR011976} for the specific purposes of upscaling and cross-validation. The locations of OzNet sites are shown as black squares in Fig. \ref{fig1}. We categorised the OzNet sites into four groups to facilitate a four-fold cross-validation. Moreover, we chose two subsets where OzNet sites have a relatively dense distribution, referred to as the cluster A and cluster B (Fig. \ref{fig1}). We also employed cluster A and B reciprocally for a cross-cluster validation. Specifically, sites within cluster A were utilised for training and predicting within cluster B, and vice versa.

\subsection{Geospatial predictors for upscaling}
We collected a diverse set of geospatial variables to serve as predictors in the upscaling process (Fig. \ref{fig2}). These predictors encompass indices and surface temperature data from the MODerate Resolution Imaging Spectroradiometer (MODIS) \cite{SCHAAF2015, WAN201436} and Landsat-resolution products \cite{5422912, GUERSCHMAN2022127318}, climatic variables from the ANUClimate 2.0 \cite{HUTCHINSON2021}, Smoothed Digital Elevation Model (DEM-S) \cite{GA2015}, and soil data from the Soil and Landscape Grid Australia \cite{MALONE2021a, MALONE2021b}.

\section{Methodology}
\begin{figure}
    \centering
    \includegraphics[width=0.35\textwidth]{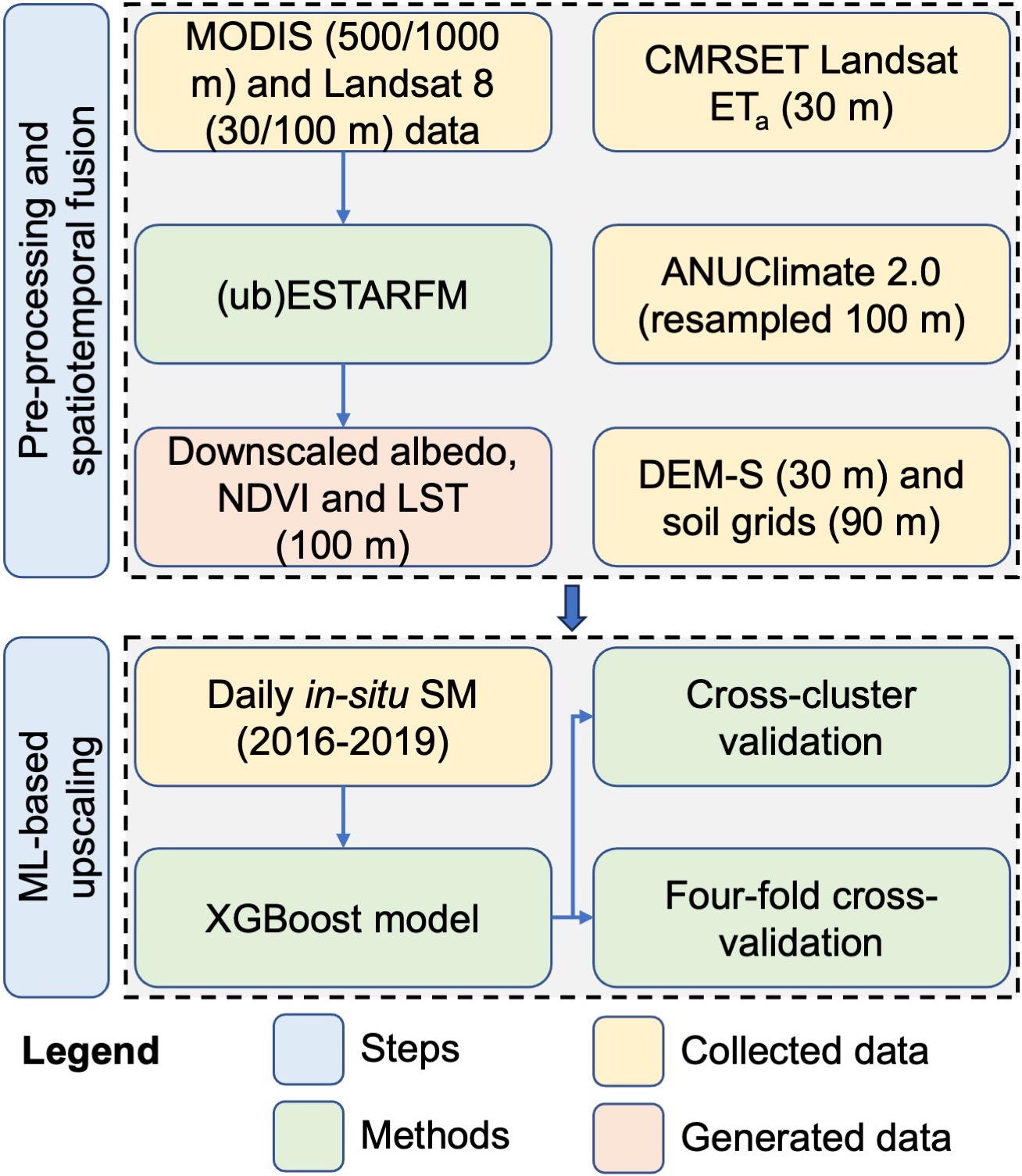}
    \caption{The experimental design herein.}
    \label{fig2}
\end{figure}

Fig. \ref{fig2} presents the experimental design employed herein, consisting of two key steps. Firstly, we utilised a nearest neighbour method to resample all predictors to 0.01° (about 100 m) resolution grids under the World Geodetic System 1984 (WGS84) datum. Additionally, we performed spatiotemporal fusion between MODIS and Landsat data to downscale MODIS albedo, NDVI, and LST to a 100 m resolution, which is more comparable to the agricultural foorprints. Secondly, we conducted machine learning (ML)-based upscaling using in-situ SM between 2016-2019 as the response, with the collected geospatial data serving as predictors. To thoroughly assess the performance of upscaling, a four-fold cross-validation and a cross-cluster validation were carried out, evaluating the transferability of the upscaling approach while accounting for regional nuances. 

\begin{figure*}[!ht]
    \centering
    \includegraphics[width=0.85\textwidth]{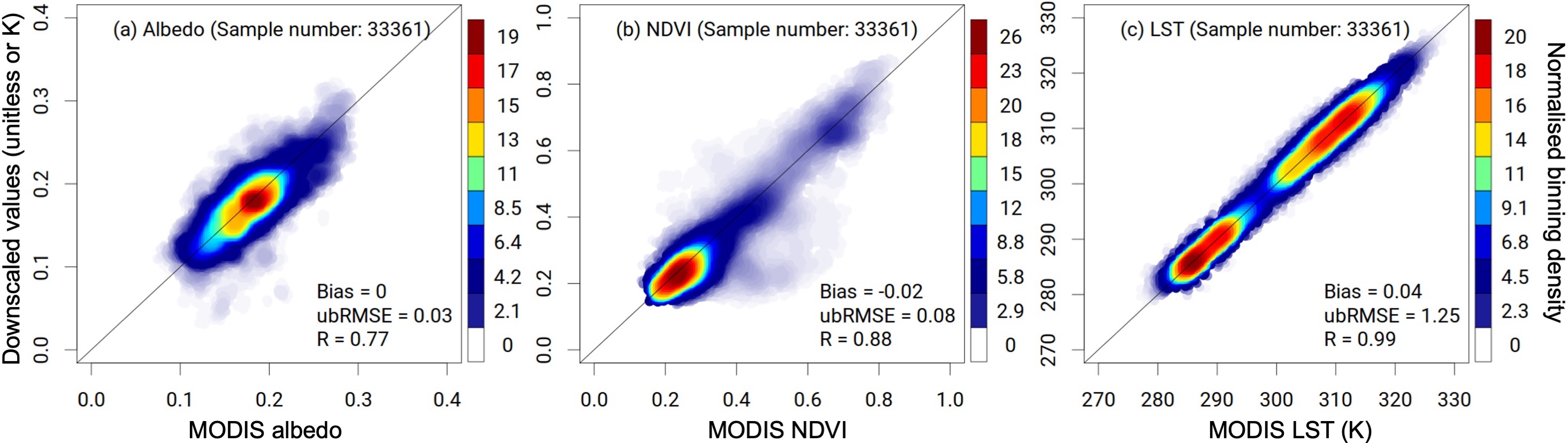}
    \caption{Density scatterplots of (a) albedo, (b) NDVI and (c) LST between 01/Jan/2016 and 31/Dec/2019 (excluding training dates of spatiotemporal fusion). The density of scatters was calculated using a binning approach.}
    \label{fig3}
\end{figure*}

\subsection{Spatiotemporal fusion}
The Enhanced Spatial and Temporal Adaptive Reflectance Fusion Model (ESTARFM) \cite{ZHU20102610} has been extensively applied to generate fine-resolution surface reflectance \cite{ZHU20102610, EMELYANOVA2013193}. It has an exceptional performance in fusing surface reflectance, while an unbiased variant (ubESTARFM) \cite{YU2023113784} has demonstrated more effectiveness in fusing land surface temperature (LST) data. Hence, we implemented ESTARFM to fuse MODIS and Landsat surface reflectance data (i.e., albedo and NDVI), and ubESTARFM to fuse MODIS and Landsat LST data, to generate daily 100 m estimates of these predictors.

\subsection{ML-based upscaling}
We employed the eXtreme Gradient Boosting (XGBoost) \cite{Friedman2001XGBoost} model to perform the upscaling of in-situ SM. ML models have showcased capability in establishing robust regression relationships between SM estimates and geospatial predictors across different scales. However, the majority of ML-based studies focused on downscaling coarse-resolution SM grids to finer resolutions \cite{im2016downscaling, ABOWARDA2021112301, modsim2021}, while fewer attempts have been made to explore the upscaling space \cite{7907192, ijgi6050130}.

\section{Results and discussion}
\subsection{Evaluations of downscaled predictors}
Fig. \ref{fig3} depicts scatterplots illustrating the downscaled performance of (a) albedo, (b) NDVI, and (c) LST in comparison to MODIS data during 01/Jan/2016 - 31/Dec/2019, encompassing 33,361 samples and using a normalised binning density. The downscaled albedo exhibited a bias of 0, ubRMSE of 0.03, and R of 0.77, with the majority of samples clustering around the value of 0.2. For the downscaled NDVI, a slight negative bias of -0.02 was observed, with an ubRMSE of 0.08 and an R of 0.88. Most NDVI values were concentrated between 0.2 and 0.3. The downscaled LST had a bias of 0.04 K, ubRMSE of 1.25 K and R of 0.99. The predominant LST concentration occurred at approximately 285 K, with a secondary concentration between 305 and 315 K. 

\begin{figure}
    \centering
    \includegraphics[width=0.45\textwidth]{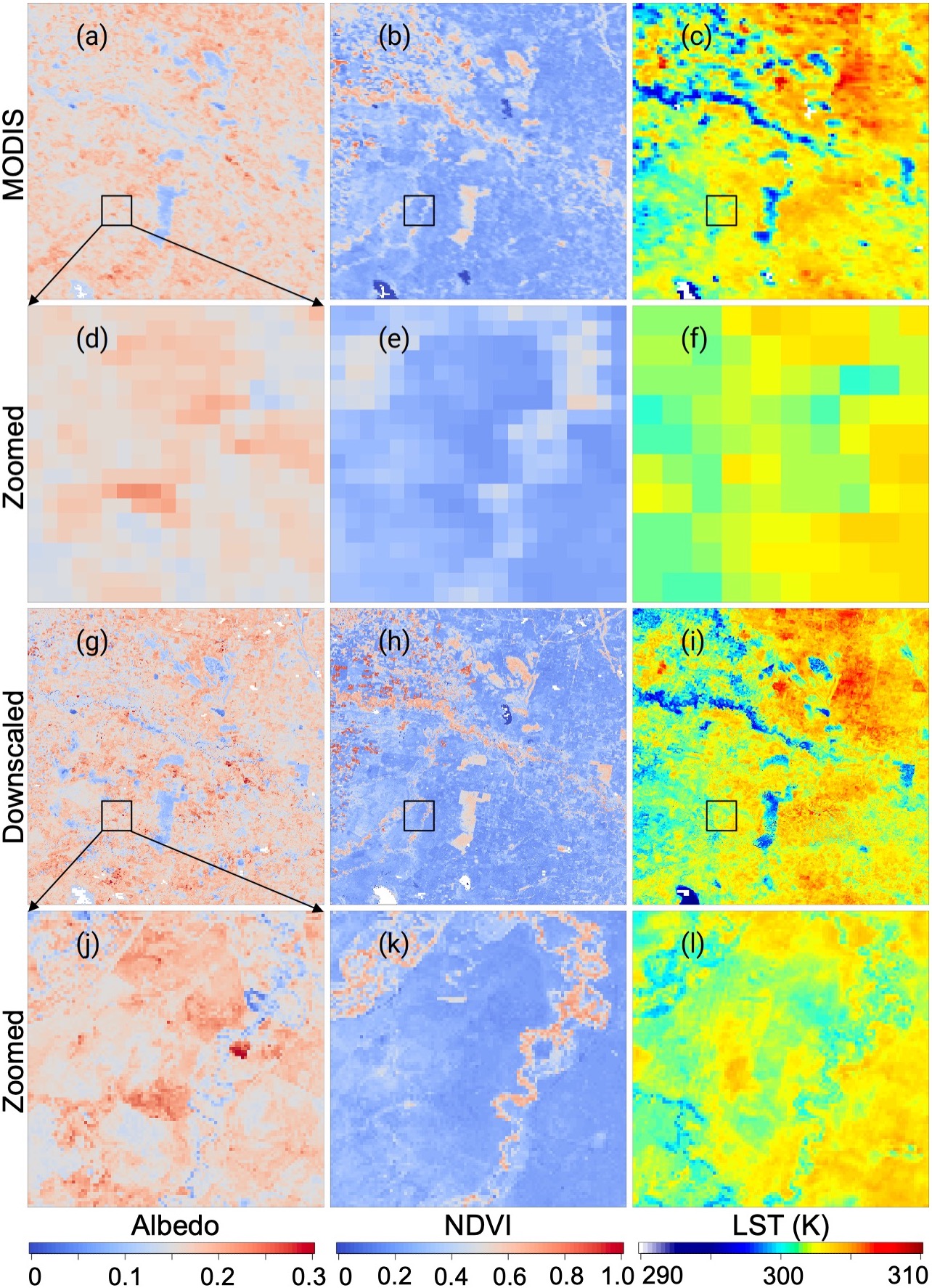}
    \caption{Spatial comparison between MODIS and downscaled predictors on 02/Apr/2017. (a-c) are MODIS albedo, NDVI and LST for the study area, respectively; (d-f) are zoomed windows of MODIS predictors spanning 146.25 to 146.35 °E and -34.95 to -35.05 °N; (g-i) are the downscaled albedo, NDVI and LST for the study area, respectively; and (j-l) are zoomed windows of downscaled predictors covering the same area with (d-f).}
    \label{fig4}
\end{figure}

Fig. \ref{fig4} provides an illustrative example on 02/Apr/2017 showcasing the spatial comparison between MODIS data and downscaled predictors. Fig. 4 (g-i) present downscaled patterns of predictors across the entire study area, demonstrating consistency with MODIS data but exhibiting sharpened features in comparison (Fig. 4 a-c). Upon closer examination of a zoomed area (Fig. 4 j-l), the downscaled predictors revealed enhanced details that are more visually discernible compared to the MODIS data (Fig. 4 d-f). In general, the spatiotemporal fusion enabled a better capture of spatial details of predictors at the field scale, offering improved representation of the agricultural landscapes.

\subsection{Evaluations of upscaled in-situ SM}
\begin{figure}
    \includegraphics[width=0.5\textwidth]{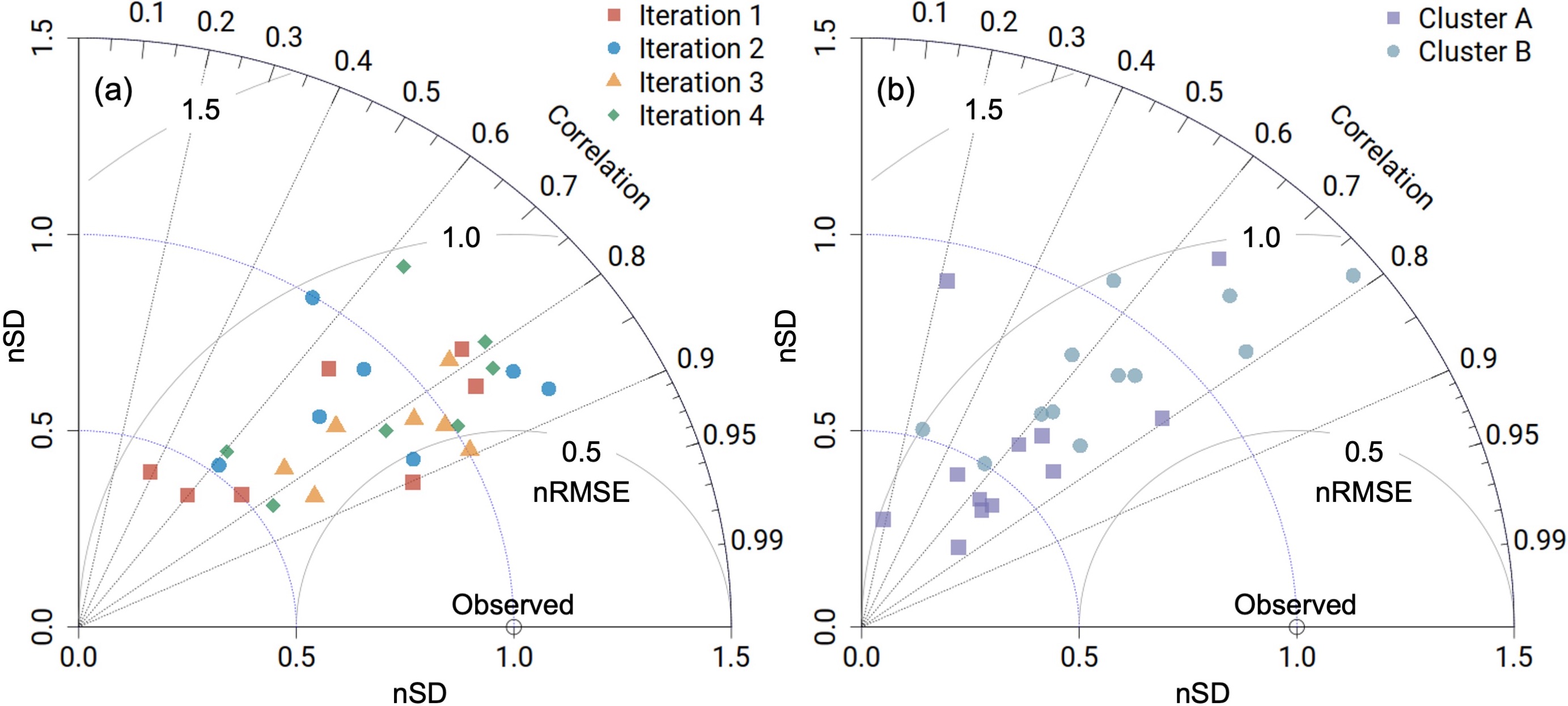}
    \caption{Normalised Taylor diagrams of upscaled SM using XGBoost during 2016-2019 for (a) four-fold cross-validation and (b) cross-cluster validation.}
    \label{fig5}
\end{figure}

Fig. \ref{fig5} presents two normalised Taylor diagrams illustrating the performance of the XGBoost model in upscaling SM using (a) four-fold cross-validation and (b) cross-cluster validation. The distribution of markers from each fold suggested variability in performance metrics across the four folds using the XGBoost model, which consistently demonstrated a correlation performance falling within the range of 0.6 to 0.9 (Fig. \ref{fig5} a). Similarly, the distribution of markers from each cluster also indicated variability in performance metrics, with the majority of markers demonstrating a correlation within the range of 0.6 to 0.8 (Fig. \ref{fig5} b). In summary, these diagrams underscore the potential transferability of the upscaling approach, showcasing its ability to maintain consistent performance across different folds and spatial clusters.

\begin{figure}
    \includegraphics[width=0.5\textwidth]{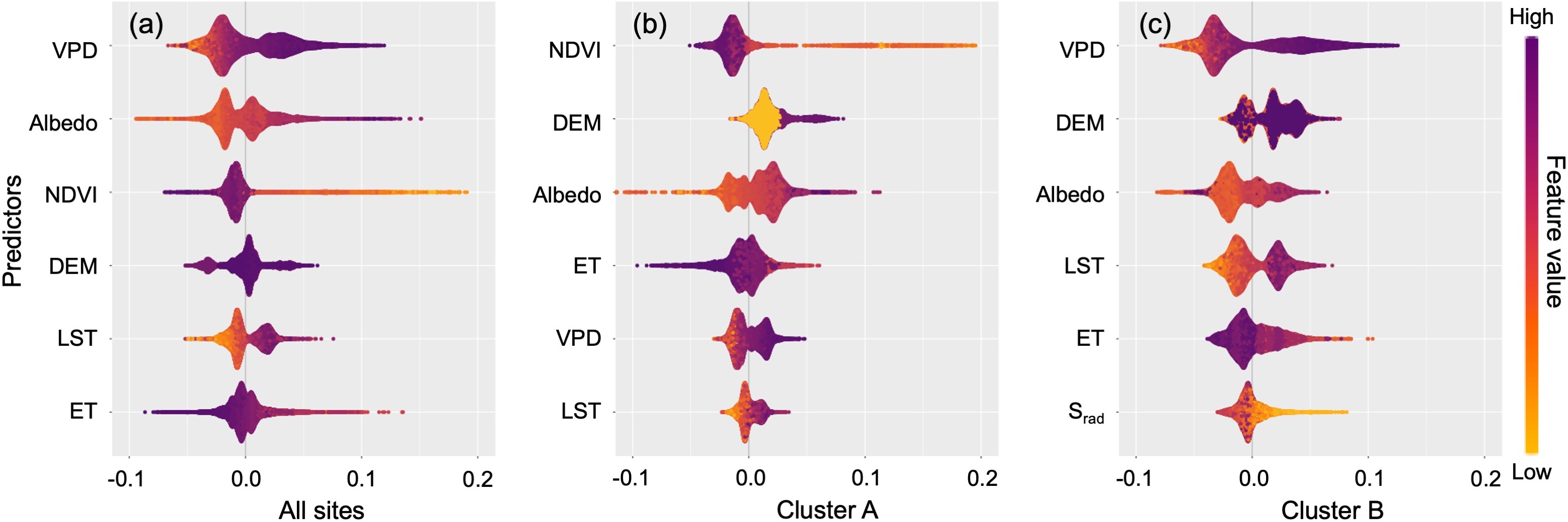}
    \caption{SHAP values of top 6 predictors of the XGBoost model using violin plots during 2016-2019 for (a) all sites; (b) cluster A; and (c) cluster B.}
    \label{fig6}
\end{figure}

Fig. \ref{fig6} depicts the SHapley Additive exPlanations (SHAP) values of the top six predictors of the XGBoost model using violin plots from 2016 to 2019 for (a) all sites; (b) cluster A; and (c) cluster B. For all sites, the order of predictors was vapour pressure deficit (VPD), albedo, NDVI, DEM, LST, and Evapotranspiration (ET). In cluster A, the order slightly changed, with NDVI at the top followed by DEM; while in cluster B, a new predictor, the incoming solar radiation (S\textsubscript{rad}), was introduced and was ranked as the 6th. The SHAP values in all three segments ranged from -0.1 to 0.2, with the majority concentrated between -0.05 and 0.05. In general, while the top predictors within the two clusters exhibited similarity, their respective rankings differed. This discrepancy suggests the model’s reliance on regional land cover and landscapes in shaping the importance and impact of specific predictors.

\begin{figure}
    \centering
    \includegraphics[width=0.45\textwidth]{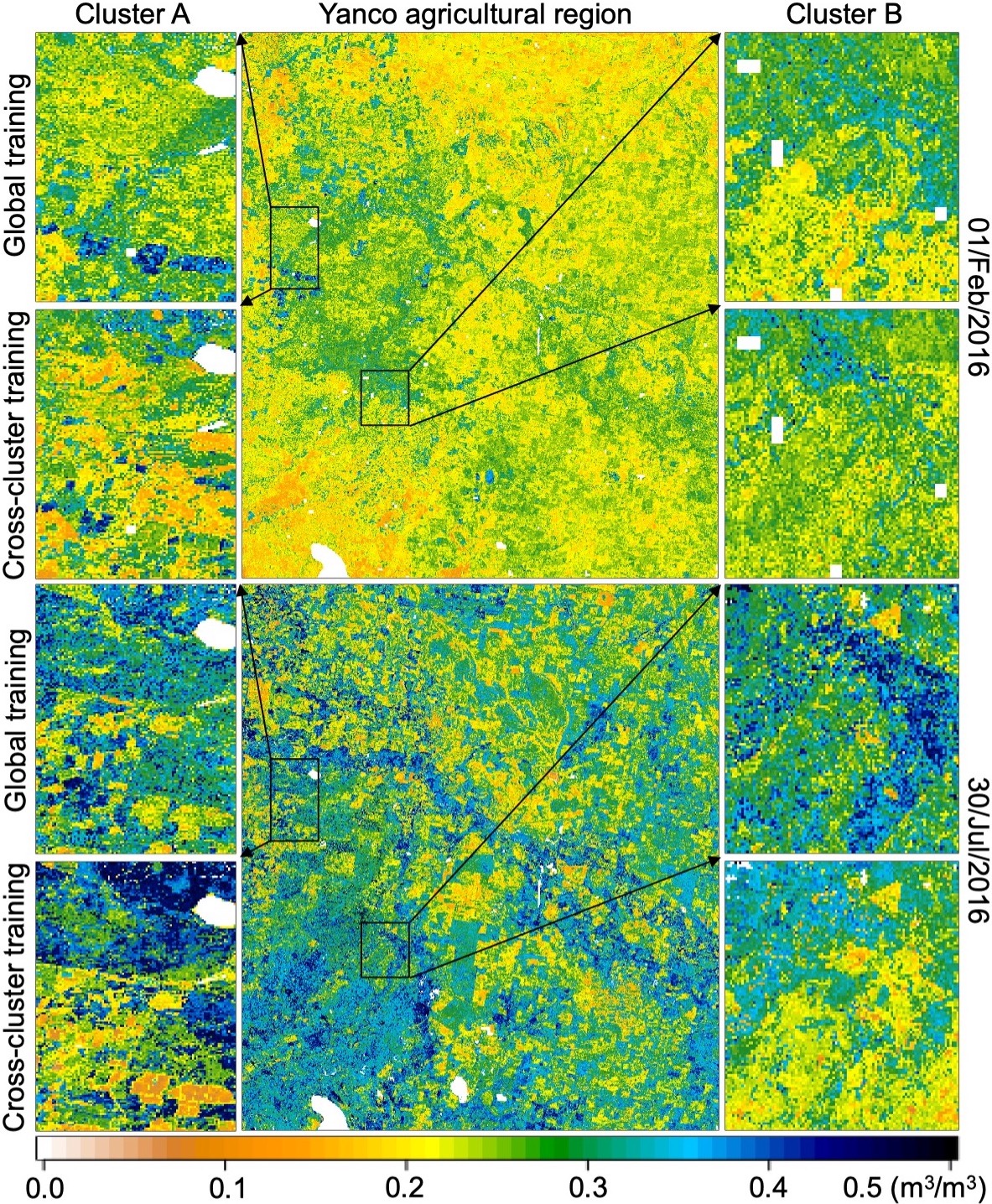}
    \caption{Spatial examples of upscaled SM using XGBoost on 01/Feb/2016 and 30/Jul/2016 for the Yanco agricultural region, zoomed areas based on global training, and zoomed areas based on cross-cluster training, respectively.}
    \label{fig7}
\end{figure}

Fig. \ref{fig7} illustrates the spatial distribution of upscaled SM utilising XGBoost in the study area, with zoomed areas focusing on two clusters at two distinct dates: 01/Feb/2016 (austral summer) and 30/Jul/2016 (austral winter). Each cluster was further compared using two different training strategies, including the global training and cross-cluster training. Notably, the cross-cluster training strategy revealed more pronounced variations in SM compared to the global training strategy. This visualisation provides insights into the effectiveness of different training strategies in upscaling SM to a 100 m resolution. The observed variations in SM across the landscape underscore the importance of considerations of regional nuances in training strategies for accurate and context-specific predictions.

\section{Conclusion}

In this study, we presented an upscaling approach that integrates spatiotemporal fusion with machine learning to extrapolate point-scale SM measurements from 28 OzNet in-situ sites to a 100 m resolution. The effectiveness of the proposed approach was assessed through multiple iterations of validation, revealing its capability to map the spatial variability of SM using sparsely distributed in-situ data, as demonstrated by the correlation performance ranging between 0.6 and 0.9 and visualised comparisons. This approach provides a pathway to extrapolate point measurements of SM to a spatial scale more comparable to climatic model grids and satellite retrievals. Future investigations should consider incorporating independent data (e.g., field campaign data) for further assessment.

\bibliography{refs}

\begin{thebibliography}{10}

\bibitem{SENEVIRATNE2010125}
Sonia~I. Seneviratne, Thierry Corti, Edouard~L. Davin, Martin Hirschi, Eric~B. Jaeger, Irene Lehner, Boris Orlowsky, and Adriaan~J. Teuling,
\newblock ``Investigating soil moisture–climate interactions in a changing climate: A review,''
\newblock {\em Earth-Science Reviews}, vol. 99, no. 3, pp. 125--161, 2010.

\bibitem{rs10122038}
Gianpaolo Balsamo, Anna Agusti-Panareda, Clement Albergel, Gabriele Arduini, Anton Beljaars, Jean Bidlot, Eleanor Blyth, Nicolas Bousserez, Souhail Boussetta, Andy Brown, Roberto Buizza, Carlo Buontempo, Frédéric Chevallier, Margarita Choulga, Hannah Cloke, Meghan~F. Cronin, Mohamed Dahoui, Patricia De~Rosnay, Paul~A. Dirmeyer, Matthias Drusch, Emanuel Dutra, Michael~B. Ek, Pierre Gentine, Helene Hewitt, Sarah~P.E. Keeley, Yann Kerr, Sujay Kumar, Cristina Lupu, Jean-François Mahfouf, Joe McNorton, Susanne Mecklenburg, Kristian Mogensen, Joaquín Muñoz-Sabater, Rene Orth, Florence Rabier, Rolf Reichle, Ben Ruston, Florian Pappenberger, Irina Sandu, Sonia~I. Seneviratne, Steffen Tietsche, Isabel~F. Trigo, Remko Uijlenhoet, Nils Wedi, R.~Iestyn Woolway, and Xubin Zeng,
\newblock ``Satellite and in situ observations for advancing global earth surface modelling: A review,''
\newblock {\em Remote Sensing}, vol. 10, no. 12, 2018.

\bibitem{DONG2020111756}
Jianzhi Dong, Wade~T. Crow, Kenneth~J. Tobin, Michael~H. Cosh, David~D. Bosch, Patrick~J. Starks, Mark Seyfried, and Chandra~Holifield Collins,
\newblock ``Comparison of microwave remote sensing and land surface modeling for surface soil moisture climatology estimation,''
\newblock {\em Remote Sensing of Environment}, vol. 242, pp. 111756, 2020.

\bibitem{PENG2021112162}
Jian Peng, Clement Albergel, Anna Balenzano, Luca Brocca, Oliver Cartus, Michael~H. Cosh, Wade~T. Crow, Katarzyna Dabrowska-Zielinska, Simon Dadson, Malcolm~W.J. Davidson, Patricia {de Rosnay}, Wouter Dorigo, Alexander Gruber, Stefan Hagemann, Martin Hirschi, Yann~H. Kerr, Francesco Lovergine, Miguel~D. Mahecha, Philip Marzahn, Francesco Mattia, Jan~Pawel Musial, Swantje Preuschmann, Rolf~H. Reichle, Giuseppe Satalino, Martyn Silgram, Peter~M. {van Bodegom}, Niko~E.C. Verhoest, Wolfgang Wagner, Jeffrey~P. Walker, Urs Wegmüller, and Alexander Loew,
\newblock ``A roadmap for high-resolution satellite soil moisture applications – confronting product characteristics with user requirements,''
\newblock {\em Remote Sensing of Environment}, vol. 252, pp. 112162, 2021.

\bibitem{reichle2004global}
Rolf~H Reichle, Randal~D Koster, Jiarui Dong, and Aaron~A Berg,
\newblock ``Global soil moisture from satellite observations, land surface models, and ground data: Implications for data assimilation,''
\newblock {\em Journal of Hydrometeorology}, vol. 5, no. 3, pp. 430--442, 2004.

\bibitem{koster2009nature}
Randal~D Koster, Zhichang Guo, Rongqian Yang, Paul~A Dirmeyer, Kenneth Mitchell, and Michael~J Puma,
\newblock ``On the nature of soil moisture in land surface models,''
\newblock {\em Journal of Climate}, vol. 22, no. 16, pp. 4322--4335, 2009.

\bibitem{dirmeyer2004comparison}
Paul~A Dirmeyer, Zhichang Guo, and Xiang Gao,
\newblock ``Comparison, validation, and transferability of eight multiyear global soil wetness products,''
\newblock {\em Journal of Hydrometeorology}, vol. 5, no. 6, pp. 1011--1033, 2004.

\bibitem{CHEN2020125054}
Yong Chen and Huiling Yuan,
\newblock ``Evaluation of nine sub-daily soil moisture model products over china using high-resolution in situ observations,''
\newblock {\em Journal of Hydrology}, vol. 588, pp. 125054, 2020.

\bibitem{https://doi.org/10.2136/vzj2010.0139}
Steffen Zacharias, Heye Bogena, Luis Samaniego, Matthias Mauder, Roland Fuß, Thomas Pütz, Mark Frenzel, Mike Schwank, Cornelia Baessler, Klaus Butterbach-Bahl, Oliver Bens, Erik Borg, Achim Brauer, Peter Dietrich, Irena Hajnsek, Gerhard Helle, Ralf Kiese, Harald Kunstmann, Stefan Klotz, Jean~Charles Munch, Hans Papen, Eckart Priesack, Hans~Peter Schmid, Rainer Steinbrecher, Ulrike Rosenbaum, Georg Teutsch, and Harry Vereecken,
\newblock ``A network of terrestrial environmental observatories in germany,''
\newblock {\em Vadose Zone Journal}, vol. 10, no. 3, pp. 955--973, 2011.

\bibitem{https://doi.org/10.1029/2012WR011976}
A.~B. Smith, J.~P. Walker, A.~W. Western, R.~I. Young, K.~M. Ellett, R.~C. Pipunic, R.~B. Grayson, L.~Siriwardena, F.~H.~S. Chiew, and H.~Richter,
\newblock ``The murrumbidgee soil moisture monitoring network data set,''
\newblock {\em Water Resources Research}, vol. 48, no. 7, 2012.

\bibitem{https://doi.org/10.1029/2011EO170001}
Wouter Dorigo, Peter van Oevelen, Wolfgang Wagner, Matthias Drusch, Susanne Mecklenburg, Alan Robock, and Thomas Jackson,
\newblock ``A new international network for in situ soil moisture data,''
\newblock {\em Eos, Transactions American Geophysical Union}, vol. 92, no. 17, pp. 141--142, 2011.

\bibitem{BROCCA2007356}
L.~Brocca, R.~Morbidelli, F.~Melone, and T.~Moramarco,
\newblock ``Soil moisture spatial variability in experimental areas of central italy,''
\newblock {\em Journal of Hydrology}, vol. 333, no. 2, pp. 356--373, 2007.

\bibitem{https://doi.org/10.1029/2011RG000372}
Wade~T. Crow, Aaron~A. Berg, Michael~H. Cosh, Alexander Loew, Binayak~P. Mohanty, Rocco Panciera, Patricia de~Rosnay, Dongryeol Ryu, and Jeffrey~P. Walker,
\newblock ``Upscaling sparse ground-based soil moisture observations for the validation of coarse-resolution satellite soil moisture products,''
\newblock {\em Reviews of Geophysics}, vol. 50, no. 2, 2012.

\bibitem{SABAGHY2020111586}
Sabah Sabaghy, Jeffrey~P. Walker, Luigi~J. Renzullo, Ruzbeh Akbar, Steven Chan, Julian Chaubell, Narendra Das, R.~Scott Dunbar, Dara Entekhabi, Anouk Gevaert, Thomas~J. Jackson, Alexander Loew, Olivier Merlin, Mahta Moghaddam, Jian Peng, Jinzheng Peng, Jeffrey Piepmeier, Christoph Rüdiger, Vivien Stefan, Xiaoling Wu, Nan Ye, and Simon Yueh,
\newblock ``Comprehensive analysis of alternative downscaled soil moisture products,''
\newblock {\em Remote Sensing of Environment}, vol. 239, pp. 111586, 2020.

\bibitem{SCHAAF2015}
Crystal Schaaf and Zhuosen Wang,
\newblock ``Mcd43a4 modis/terra+aqua nadir brdf-adjusted reflectance daily l3 global - 500m [dataset],''
\newblock {\em NASA LP DAAC}.

\bibitem{WAN201436}
Zhengming Wan,
\newblock ``New refinements and validation of the collection-6 modis land-surface temperature/emissivity product,''
\newblock {\em Remote Sensing of Environment}, vol. 140, pp. 36--45, 2014.

\bibitem{5422912}
Fuqin Li, David L.~B. Jupp, Shanti Reddy, Leo Lymburner, Norman Mueller, Peter Tan, and Anisul Islam,
\newblock ``An evaluation of the use of atmospheric and brdf correction to standardize landsat data,''
\newblock {\em IEEE Journal of Selected Topics in Applied Earth Observations and Remote Sensing}, vol. 3, no. 3, pp. 257--270, 2010.

\bibitem{GUERSCHMAN2022127318}
Juan~P. Guerschman, Tim~R. McVicar, Jamie Vleeshower, Thomas~G. {Van Niel}, Jorge~L. Peña-Arancibia, and Yun Chen,
\newblock ``Estimating actual evapotranspiration at field-to-continent scales by calibrating the cmrset algorithm with modis, viirs, landsat and sentinel-2 data,''
\newblock {\em Journal of Hydrology}, vol. 605, pp. 127318, 2022.

\bibitem{HUTCHINSON2021}
M.~F. Hutchinson, T.~Xu, J.~L. Kesteven, I.~J. Marang, and B.~J. Evans,
\newblock ``Anuclimate v2.0 [dataset],''
\newblock {\em NCI Australia, Canberra}, 2021.

\bibitem{GA2015}
GA,
\newblock ``Digital elevation model (dem) of australia derived from lidar 5 metre grid,''
\newblock {\em Geoscience Australia, Canberra}, 2015.

\bibitem{MALONE2021a}
Brendan Malone and Ross Searle,
\newblock ``Updating the australian digital soil texture mapping (part 1): re-calibration of field soil texture class centroids and description of a field soil texture conversion algorithm,''
\newblock {\em Soil Research}, vol. 59, pp. 419--434, 2021.

\bibitem{MALONE2021b}
Brendan Malone and Ross Searle,
\newblock ``Updating the australian digital soil texture mapping (part 2): spatial modelling of merged field and lab measurements,''
\newblock {\em Soil Research}, vol. 59, pp. 435--451, 2021.

\bibitem{ZHU20102610}
Xiaolin Zhu, Jin Chen, Feng Gao, Xuehong Chen, and Jeffrey~G. Masek,
\newblock ``An enhanced spatial and temporal adaptive reflectance fusion model for complex heterogeneous regions,''
\newblock {\em Remote Sensing of Environment}, vol. 114, no. 11, pp. 2610--2623, 2010.

\bibitem{EMELYANOVA2013193}
Irina~V. Emelyanova, Tim~R. McVicar, Thomas~G. {Van Niel}, Ling~Tao Li, and Albert~I.J.M. {van Dijk},
\newblock ``Assessing the accuracy of blending landsat–modis surface reflectances in two landscapes with contrasting spatial and temporal dynamics: A framework for algorithm selection,''
\newblock {\em Remote Sensing of Environment}, vol. 133, pp. 193--209, 2013.

\bibitem{YU2023113784}
Yi~Yu, Luigi~J. Renzullo, Tim~R. McVicar, Brendan~P. Malone, and Siyuan Tian,
\newblock ``Generating daily 100 m resolution land surface temperature estimates continentally using an unbiased spatiotemporal fusion approach,''
\newblock {\em Remote Sensing of Environment}, vol. 297, pp. 113784, 2023.

\bibitem{Friedman2001XGBoost}
Jerome~H. Friedman,
\newblock ``Greedy function approximation: A gradient boosting machine,''
\newblock {\em The Annals of Statistics}, vol. 29, no. 5, pp. 1189--1232, 2001.

\bibitem{im2016downscaling}
Jungho Im, Seonyoung Park, Jinyoung Rhee, Jongjin Baik, and Minha Choi,
\newblock ``Downscaling of amsr-e soil moisture with modis products using machine learning approaches,''
\newblock {\em Environmental Earth Sciences}, vol. 75, pp. 1--19, 2016.

\bibitem{ABOWARDA2021112301}
Ahmed~Samir Abowarda, Liangliang Bai, Caijin Zhang, Di~Long, Xueying Li, Qi~Huang, and Zhangli Sun,
\newblock ``Generating surface soil moisture at 30 m spatial resolution using both data fusion and machine learning toward better water resources management at the field scale,''
\newblock {\em Remote Sensing of Environment}, vol. 255, pp. 112301, 2021.

\bibitem{modsim2021}
Yi~Yu, Luigi~J. Renzullo, and Siyuan Tian,
\newblock ``Continental scale downscaling of awra-l analysed soil moisture using random forest regression,''
\newblock {\em MODSIM2021, 24th International Congress on Modelling and Simulation, Sydney, Australia, 5-10 December}, 2021.

\bibitem{7907192}
Daniel Clewley, Jane~B. Whitcomb, Ruzbeh Akbar, Agnelo~R. Silva, Aaron Berg, Justin~R. Adams, Todd Caldwell, Dara Entekhabi, and Mahta Moghaddam,
\newblock ``A method for upscaling in situ soil moisture measurements to satellite footprint scale using random forests,''
\newblock {\em IEEE Journal of Selected Topics in Applied Earth Observations and Remote Sensing}, vol. 10, no. 6, pp. 2663--2673, 2017.

\bibitem{ijgi6050130}
Dongying Zhang, Wen Zhang, Wei Huang, Zhiming Hong, and Lingkui Meng,
\newblock ``Upscaling of surface soil moisture using a deep learning model with viirs rdr,''
\newblock {\em ISPRS International Journal of Geo-Information}, vol. 6, no. 5, 2017.

\end{thebibliography}
\bibliographystyle{IEEEbib}

\end{document}